\documentclass[letterpaper,times]{IONconf}
\usepackage{amsmath}
\usepackage{amssymb}

\usepackage{graphicx}
\usepackage{booktabs}
\usepackage{multirow}

\usepackage{url}

\usepackage{natbib}

\usepackage[hidelinks]{hyperref}

\title{Automatic Operation of an Articulated Dump Truck: State Estimation by Combined QZSS CLAS and Moving-Base RTK Using Multiple GNSS Receivers}

\author{
    Taro~Suzuki, \textit{Chiba~Institute~of~Technology}
    \vspace{1mm} \\%
    Shotaro~Kojima, Kazunori~Ohno, Naoto~Miyamoto, Takahiro~Suzuki, \textit{Tohoku~University}%
    \vspace{1mm} \\%
    Kimitaka~Asano, \textit{Sanyo-Technics Co., Ltd.}, Tomohiro~Komatsu, \textit{Kowatech Co., Ltd.}, Hiroto~Kakizaki, \textit{Sato Koumuten Co., Ltd.}
    }

\begin{document}

\maketitle

\section*{biography}


\biography{Taro Suzuki}{is a chief researcher at Chiba Institute of Technology, Japan. He received his B.S., M.S., and Ph.D. in Engineering from Waseda University in 2007, 2009, and 2012 respectively. From 2012 to 2014, he worked as a postdoctoral researcher at Tokyo University of Marine Science and Technology. From 2015 to 2019, he worked as an assistant professor at Waseda University. His current research interests include GNSS precise positioning in urban environments.}

\biography{Shotaro Kojima}{received BS, MS, and Ph.D. from Tohoku University in 2015, 2017, and 2021. He has been an assistant professor of New Industry Creation Hatchery Center, Tohoku
University since 2021 to 2022. He is an assistant professor of Tough Cyberphysical AI Research Center (TCPAI), Tohoku University (2023-).}

\biography{Kazunori Ohno}{received BS, MS, and Ph.D. from Tsukuba University in 1999, 2001, and 2004. He has been an assistant professor, lecturer, associate professor Tohoku University, and now professor of New Industry Creation Hatchery Center (NICHe) Tohoku University since 2021. He also serves as chair of Data Engineering Robotics and head of the Physical Division of the Tough Cyber-physical AI Research Center. His research fields are field robotics, robot intelligence, and cyber-enhanced canines.}

\biography{Naoto Miyamoto}{received BS, MS and Ph.D. degrees in electronic engineering from Tohoku University, Japan, in 2002, 2004 and 2007 respectively. He is engaged in research at Tohoku University and Doshisha University while working at Ohtake-Root Kogyo Co., Ltd.}

\biography{Takahiro Suzuki}{received BS, MS, and Dr. of Engineering from the University of Tokyo in 1993, 1995, and 1998. He has been an Associate Professor of Univ. of Tokyo since 2000 to 2010, a Director of Nagasaki Prefecture since 2010 to 2013, and a Professor of Tohoku University since 2014 to 2023. He is now a Professor of Reitaku University, Japan, since 2023.}

\biography{Kimitaka Asano}{is a president in Sanyo-Technics Co.,Ltd. He has engaged in research on the retrofit robot technologies and the autonomous driving of large dump trucks.}

\biography{Tomohiro Komatsu}{received BS and MS degrees in mechanical engineering from Tokai University in 2010 and 2014, respectively, and Ph.D. from Tohoku University in 2023. He works at Kowatech Corporation since 2014.}

\biography{Hiroto Kakizaki}{is an employee in Sato koumuten Co.,Ltd. He has engaged in research on the autonomous driving of large dump trucks and cooperation between human operators and construction machinery.}

\section*{Abstract}
Labor shortage due to the declining birth rate has become a serious problem in the construction industry, and automation of construction work is attracting attention as a solution to this problem. This paper proposes a method to realize state estimation of dump truck position, orientation and articulation angle using multiple GNSS for automatic operation of dump trucks. RTK-GNSS is commonly used for automation of construction equipment, but in mountainous areas, mobile networks often unstable, and RTK-GNSS using GNSS reference stations cannot be used. Therefore, this paper develops a state estimation method for dump trucks that does not require a GNSS reference station by using the Centimeter Level Augmentation Service (CLAS) of the Japanese Quasi-Zenith Satellite System (QZSS). Although CLAS is capable of centimeter-level position estimation, its positioning accuracy and ambiguity fix rate are lower than those of RTK-GNSS. To solve this problem, we construct a state estimation method by factor graph optimization that combines CLAS positioning and moving-base RTK-GNSS between multiple GNSS antennas. Evaluation tests under real-world environments have shown that the proposed method can estimate the state of dump trucks with the same accuracy as conventional RTK-GNSS, but does not require a GNSS reference station.

\section{Introduction}
In the construction industry, the labor shortage due to the declining birth rate and aging population has become a serious social problem. This labor shortage is expected to accelerate in the future, and immediate countermeasures are needed. Therefore, automation of construction equipment is attracting attention as a means to alleviate this labor shortage and realize labor-saving construction \cite{con1,con2,con3}. At construction sites in mountainous areas and mines, the transportation of earth and sand by dump trucks is one of the most basic and frequent tasks, and automation of this task is desired \cite{congeneral1,congeneral2}. The automation of earth moving by dump trucks is relatively simple compared to excavation by backhoes, and is considered to be the lowest hurdle for the introduction of automated systems. However, current construction companies, including small and medium-sized enterprises, face the problem that it is cost prohibitive to introduce new dump trucks capable of automated operation.

So far, we have developed an automated driving system that estimates various states of an articulated dump truck by retrofitting four GNSS receivers/antennas \cite{our1,our2,taro_AR}. Figure 1 shows an articulated dump truck equipped with four GNSS receivers/antennas. The articulated dump truck differs from a normal dump truck in that it controls the direction of travel by hydraulically turning the front section against the rear section and bending the joint connecting the front and rear. The angle sustained by this joint is called the "articulattion angle.” Our previouse paper uses RTK-GNSS with a GNSS reference station to estimate the position, attitude, and articulation angle of the dump truck based on the geometric arrangement of the four GNSS antennas. GNSS reference stations for RTK-GNSS and mobile networks for data communication are essential for this method. However, in mountainous and mining environments, there are many areas where cellular networks are not available, and the rugged terrain and large working area make stable use of local radio communications difficult. Therefore, the automatic operation of dump trucks using RTK-GNSS with GNSS reference stations has had limited applicability in actual mines and mountainous areas.

\begin{figure}[t]
    \centering
    \includegraphics[width=110mm]{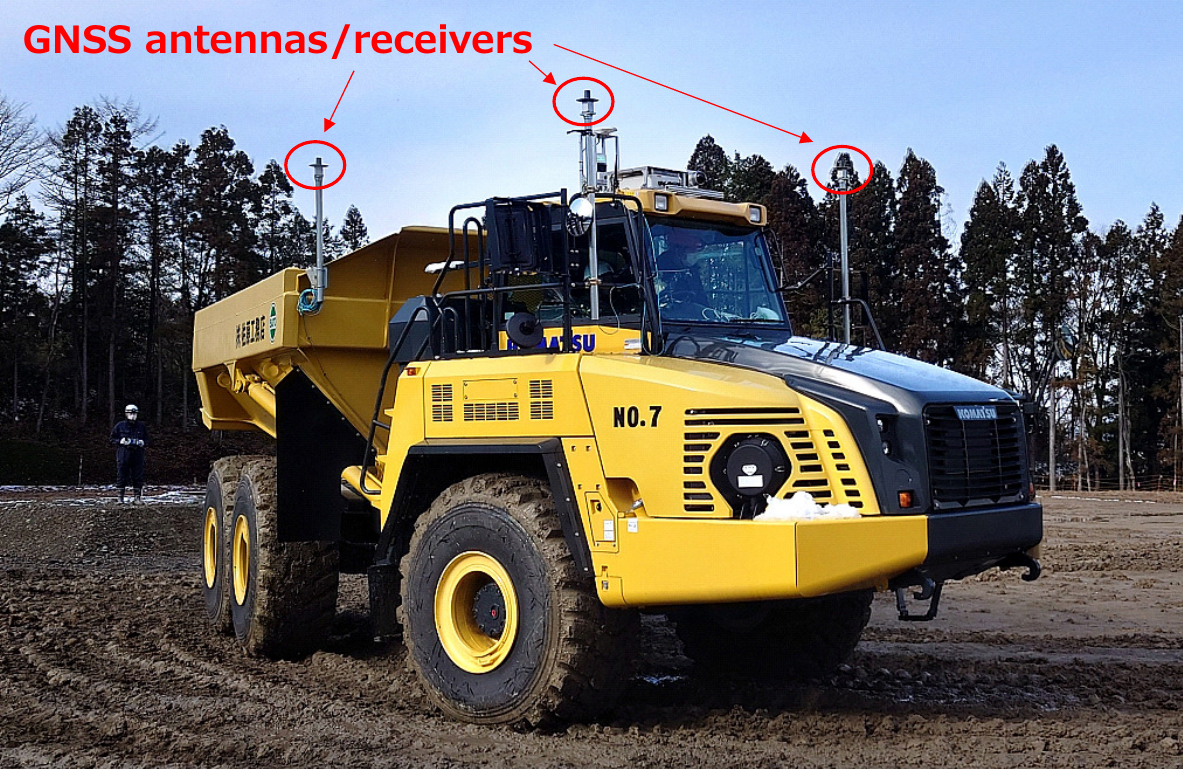} 
    \caption{Six-wheeled articulated dump truck for sediment transportation. The front and rear sections of the dump truck bend to control the direction of travel. Four GNSS receivers and antennas are installed to estimate the condition of the dump truck.}
    \label{fig:1}
\end{figure}

In this paper, we propose an automated dump truck driving system that combines the Centimeter Level Augmentation Service (CLAS) of Japanese Quasi-Zenith Satellite System (QZSS) and moving-base RTK-GNSS between multiple antennas. The use of CLAS will enable automatic operation of dump trucks without the need for ground communication and GNSS reference stations, and is expected to greatly expand the application environment for automatic operation of dump trucks.

\section{Related Works}
In some studies, dump trucks are directly operated by controlling their hydraulic system \cite{art1,art2}. However, this mechanism of operation is difficult for users because construction machinery manufacturers do not disclose vehicle information. Therefore, in this study, a robot was retrofitted to physically control the dump truck. We developed a robot named SAM, which enables the remote control of an existing hydraulic dump truck \cite{our1}. However, in order to realize automatic operation of dump trucks, it is necessary to measure various states of dump trucks.

QZSS CLAS uses a method called PPP-RTK, which transmits GNSS correction information for high-precision positioning via the QZSS L6 signal \cite{clas}. In CLAS-based positioning, the GNSS observations of a virtual reference station are generated from the correction information, and the usual RTK-GNSS algorithm, which uses the double difference in carrier phase, can be used for high-precision positioning without using a real GNSS reference station. However, communication bandwidth limitations limit the number of satellites to be corrected. CLAS has a defined horizontal accuracy of 6 cm (95\%) and vertical accuracy of 12 cm (95\%) in a stationary environment, but the horizontal accuracy deteriorates to 12 cm (95\%) and vertical accuracy to 24 cm (95\%) in a mobile environment. The time to first fix (TTFF) is 60 seconds. Therefore, the positioning accuracy and ambiguity fix rate are lower than those of conventional RTK-GNSS.

Research on CLAS-based control of robots and mobile vehicles has attracted much attention in recent years. In References \cite{clas_agri1} and  \cite{clas_agri2}, CLAS-compatible GNSS receivers and antennas are mounted on agricultural machinery, and it is shown that automatic control of agricultural machinery is possible. Reference \cite{clas_eva} also evaluates the accuracy of CLAS positioning results in urban areas and various environments on mobile vehicles. Reference \cite{taro_clas} evaluates the performance of CLAS-compatible L6 antennas and the accuracy of positioning results from mobile vehicles. As quantitative evaluations of CLAS in stationary environments, References \cite{clas_eva3} and \cite{clas_eva} report that CLAS positioning accuracy is close to specification. However, there are no examples of CLAS positioning being used for automatic operation of dump trucks, and applications using multiple CLAS-enabled GNSS receivers have not been investigated. 

When CLAS is used to estimate the state of a dump truck, the aforementioned deterioration in the accuracy of CLAS positioning, especially in estimating the orientation angle and articulation angle, is a problem. Therefore, this paper proposes a new method for estimating the state of dump trucks that overcomes the weakness of QZSS CLAS and does not use a terrestrial GNSS reference station.

\section{Methodology}
\subsection{Overview of Proposed Methodology}
The objective of this study is to develop an automatic operation system for dump trucks that does not use a GNSS reference station, using CLAS. When CLAS is used for state estimation of dump trucks, the positioning accuracy is slightly inferior to that of RTK-GNSS using GNSS reference stations, and the ambiguity correction rate is also worse, resulting in a worse state estimation accuracy. In this study, we propose a new graph optimization method-based dump truck state estimation method that combines absolute position estimation using CLAS and moving-based RTK-GNSS from multiple GNSS receivers mounted on dump trucks.

\begin{figure}[t!]
    \centering
    \includegraphics[width=110mm]{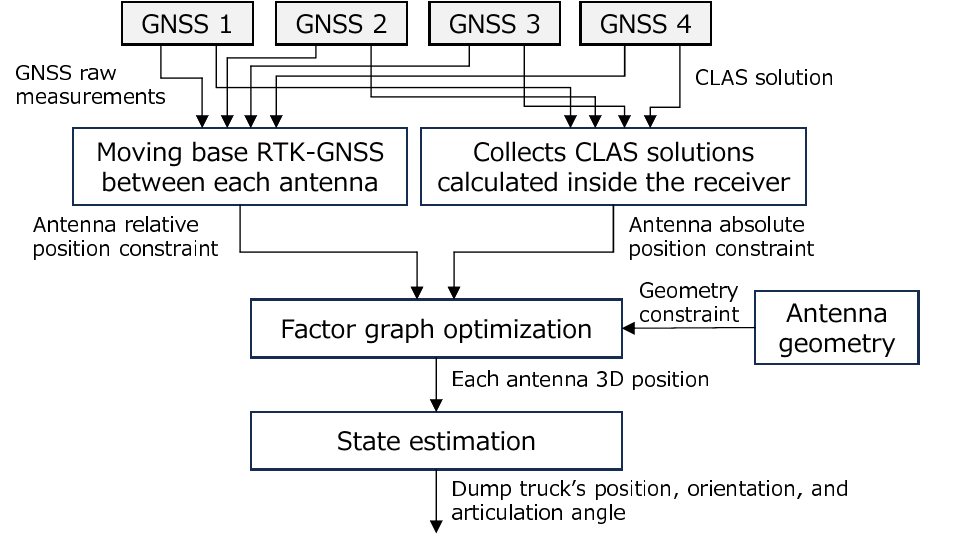} 
    \caption{An overview of the proposed system and method. Absolute positions of antennas are estimated by CLAS positioning, and relative positions between antennas are estimated by moving-base RTK-GNSS to estimate orientation and articulation angles of dump trucks.}
    \label{fig:2}
\end{figure}

Figure 2 shows an overview of the proposed method. The basic idea of the proposed method is to (1) estimate the orientation and articulation angle of a dump truck with high accuracy by using relative position constraints between antennas based on moving-base RTK-GNSS between multiple antennas, and (2) improve the availability and accuracy of dump truck position estimation by combining the absolute position of each antenna estimated by CLAS with factor graph optimization to improve the accuracy of dump truck location estimation. Compared to the relative positions between antennas obtained from CLAS, the moving-base RTK can be used to estimate the baseline vectors between multiple antennas with high accuracy. This relative position constraint between antennas improves the accuracy of orientation and articulation angle estimation. Absolute position estimation improves the accuracy of CLAS positioning by combining CLAS solutions from multiple GNSS and enables highly accurate position estimation of dump trucks even when ambiguity-fixed solutions are not available for some antennas.

\subsection{System Configuration}
Figure 3 shows the developed antenna-integrated CLAS-compliant GNSS receiver using u-blox ZED-F9R for GNSS reception and u-blox ZED-D9C for CLAS reception and decoding, connected to a single board computer (Rodxa ROCK Pi S) with PoE functionality. The antenna is a Tallysman HC976 capable of receiving L6 signals. The power supply and communication are provided via an Ethernet cable, and the antenna positioning results from CLAS and the raw GNSS observations (pseudorange, carrier phase, etc.) for moving-base RTK-GNSS are sent to the central processing single board computer (Raspberry Pi 4) via TCP/IP communication. Here, the satellite systems of the signals received by each GNSS receiver are limited to GPS, QZSS, and Galileo, which are CLAS correction targets, and the observation data are transmitted with a 20 Hz observation cycle, and positioning is performed using CLAS. A single-board computer for central processing estimates the orientation, articulation angle, and position of the dump truck in a 20-Hz period and uses this information to control the dump truck. 

\begin{figure}[t]
    \centering
    \includegraphics[width=130mm]{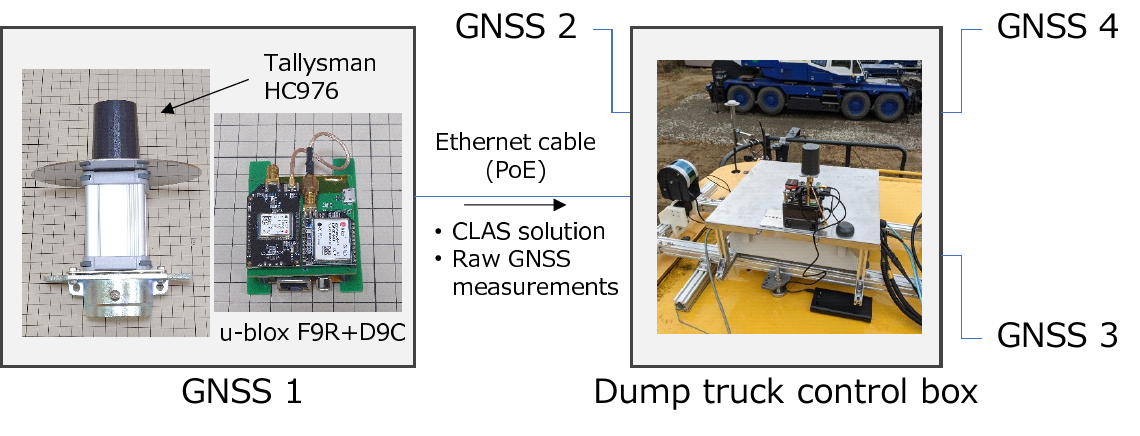} 
    \caption{The developed antenna-integrated CLAS-compliant GNSS receiver, which outputs CLAS positioning results and raw GNSS observations via TCP/IP.}
    \label{fig:3}
\end{figure}

\subsection{Factor Graph Optimization}
We use factor graph optimization to estimate the state of a dump truck. Factor graph optimization is a state estimation technique that is currently popular in the field of robotics \cite{go1}, and is also gaining attention in the field of GNSS as an alternative to traditional Kalman filter-based state estimation methods \cite{taro_RAL2022,taro_gsdc2022}. Factor graphs are graph representations in which variable nodes and factor nodes are connected by edges, with the estimated state as the variable node and the observation as the factor node. Factor acts as a constraint on the connected variable nodes and encodes an error function. The structure of the graph proposed in this research is shown in Figure 4. Let $\mathbf{x}_{i}=\left[\begin{array}{lll}x_i & y_i & z_i\end{array}\right]$ be the 3D position of the $i$-th antenna in the local east-north-up (ENU) coordinate system. The estimated state is the 3D position of the four antennas at the current time, as follows. 

\begin{equation}
    \mathbf{X}=\left[\begin{array}{llll}
    \mathbf{x}_{1} & \mathbf{x}_{2} & \mathbf{x}_{3} & \mathbf{x}_{4}
    \end{array}\right]^{T}
\end{equation}

\begin{figure}[t]
    \centering
    \includegraphics[width=100mm]{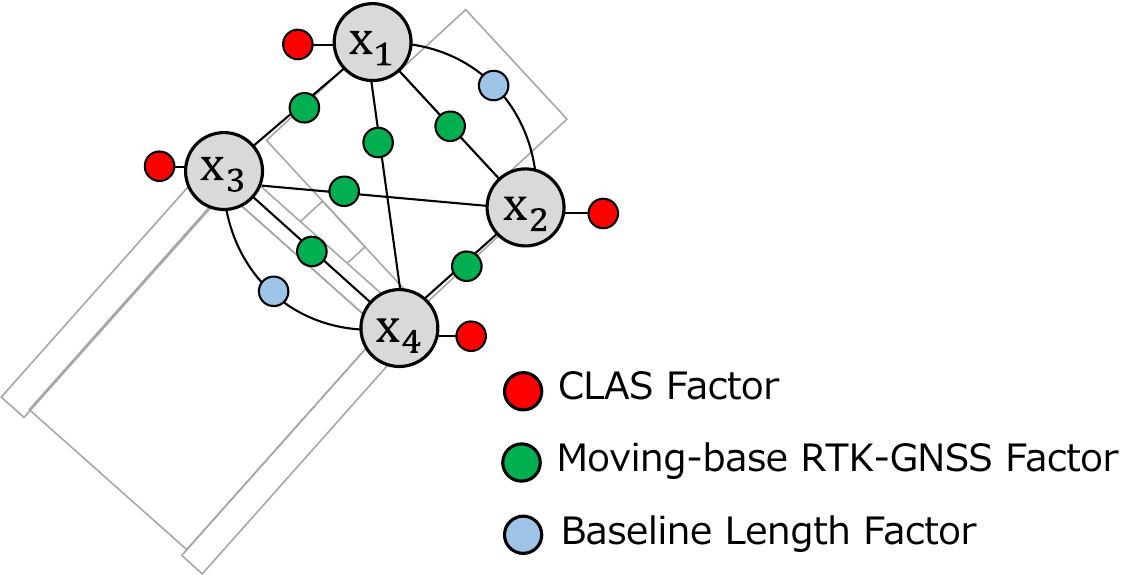} 
    \caption{The structure of the graph proposed in this research. Three factors are used: the 3D position factor of CLAS, the moving-base RTK-GNSS factor between the antennas, and the baseline length factor between the antennas.}
    \label{fig:4}
\end{figure}

In this paper, three factors are used: the 3D position factor of CLAS, the moving-base RTK-GNSS factor between the antennas, and the baseline length factor between the antennas. By integrating multiple factors to obtain the 3D position of the antennas, the state of the dump truck can be accurately estimated even when an ambiguity-fixed solution for a single antenna is not available. Since dump truck control requires real-time state estimation, each time GNSS observations are obtained, a graph as shown in Figure 4 is created and the position of each antenna is estimated by optimization. In this paper, past observations are not used to reduce computational complexity, but if the trajectory of the dump truck is to be estimated in post-processing, a large graph can be created from the state and observations at all times, and all states can be estimated in the same way.

\subsubsection{CLAS Factor}
The absolute position of the dump truck is constrained using the PPP-RTK solution from CLAS. The CLAS positioning solution is provided by the u-blox F9R, which reduces the amount of computation required. Let $\mathbf{b}_{i}$ be the CLAS solution of $i$-th angtenna converted to the local ENU coordinate system, and the error function of the CLAS factor is expressed as follows.

\begin{equation}
    \mathbf{e}_{\mathrm{clas},i}=\mathbf{x}_i-\mathbf{b}_i
\end{equation}
    
\begin{equation}
    \left\|\mathbf{e}_{\mathrm{clas},i}\right\|_{\mathbf{\Omega}_{\mathrm{clas}}}=\mathbf{e}_{\mathrm{clas},i} \Omega_\mathrm{clas} \mathbf{e}_{\mathrm{clas},i}
\end{equation}
  
\noindent where $\mathbf{\Omega}_{\mathrm{clas}}$ is the information matrix and is the inverse of the error variance of the CLAS positioning solution output by the u-blox receiver. The factor is only added to the graph if the CLAS solution status is a float or fix.

\subsubsection{Moving-Base RTK-GNSS Factor}
The baseline vectors between all antennas, obtained from moving-base RTK-GNSS between antennas, are added to the graph as relative position constraints between antennas. Here, each GNSS receiver sends raw GNSS observations (carrier phase and pseudorange) to the central processing single-board computer simultaneously with the positioning solution using CLAS, and moving-base RTK-GNSS is performed using RTKLIB \cite{rtklib} from these multi-antenna GNSS observations \cite{movingbase}. Here there are six combinations of four antennas, excluding duplicates ${}_4 C _2=6$. Relative positions are calculated from moving-base RTK-GNSS for all antenna combinations.

The moving-base RTK-GNSS factor is added to the graph as a relative constraint between antennas only when a carrier phase ambiguity fixed solution is obtained. The error function of the factor is as follows.

\begin{equation}
    \mathbf{e}_{\mathrm{mvrtk},m}=\mathbf{x}_{S^1(m)}-\mathbf{x}_{S^2(m)}-\mathbf{b}_m
\end{equation}
    
\begin{equation}
    \left\|\mathbf{e}_{\mathrm{mvrtk},m}\right\|_{\mathbf{\Omega}_{\mathrm{mvrtk}}}=\mathbf{e}_{\mathrm{mvrtk},m} \mathbf{\Omega}_\mathrm{mvrtk} \mathbf{e}_{\mathrm{mvrtk},m}
\end{equation}

\noindent where $S^1(m)$ and $S^2(m)$ represent the antenna number of the $m$-th GNSS antenna combinations. $\mathbf{\Omega}_{\mathrm{mvrtk}}$ is the inverse of the variance of the moving-base RTK-GNSS positioning solution. The moving-base RTK-GNSS factor can improve positioning accuracy in environments where CLAS positioning accuracy is degraded.

\subsubsection{Baseline Length Factor}
Since the geometric arrangement of the GNSS antennas is fixed on the dump truck, the geometric distance between the antennas obtained as prior knowledge can be used as a constraint. Here, in articulated dump trucks, the geometric arrangement of the front and rear parts of the dump truck changes during driving. Therefore, the geometric distance between the antennas in the front part (Antenna 1 and Antenna 2) and between the antennas in the rear part (Antenna 3 and Antenna 4) is used as a constraint condition. If the 3D distances between the two antennas are $L_{12}$ and $L_{34}$, respectively, the baseline distance factor is expressed as follows.

\begin{equation}
    e_{\mathrm{ant},12}=\left\|\mathbf{x}_\mathrm{2}-\mathbf{x}_\mathrm{1}\right\|-L_{12}, e_{\mathrm{ant},34}=\left\|\mathbf{x}_\mathrm{4}-\mathbf{x}_\mathrm{3}\right\|-L_{34}
\end{equation}
  
\begin{equation}
    \left\|e_{\mathrm{ant}}\right\|_{\Omega_{\mathrm{ant}}}=e_{\mathrm{ant},12} \Omega_\mathrm{ant} e_{\mathrm{ant},12}+e_{\mathrm{ant},34} \Omega_\mathrm{ant} e_{\mathrm{ant},34}
\end{equation}

\noindent where $\Omega_\mathrm{ant}$ is an information matrix about the geometric distance between antennas, and since the geometric distance between antennas is not expected to vary, a very small variance ($0.01^2$ $m^2$ in this paper) is used as the variance.

\subsubsection{Optimization}
The final objective function to be optimized is as follows.

\begin{equation}
    \widehat{\mathbf{X}}=\underset{\mathbf{X}}{\operatorname{argmin}} 
    \sum_{i} \left\|\mathbf{e}_{\mathrm{clas}, i}\right\|_{\mathbf{\Omega}_{\mathrm{clas},i}}^{2} +
    \sum_{m}\left\|\mathbf{e}_{\mathrm{mvrtk}, m}\right\|_{\mathbf{\Omega}_{\mathrm{mvrtk, m}}}^{2} +
    \left\|e_{\mathrm{ant}}\right\|_{\Omega_{ \mathrm{ant}}}^{2}
\end{equation}

\noindent where Gauss-Newton optimization is used as the optimization method \cite{gtsam}. In addition, the M-estimator is used for the CLAS factor to reduce the influence of outliers such as wrong ambiguity fixed solutions. The Huber function is used as the kernel of the M-estimator.

\subsection{Position, Orientation, and Articulation Angle Estimation}

\begin{figure}[t!]
    \centering
    \includegraphics[width=50mm]{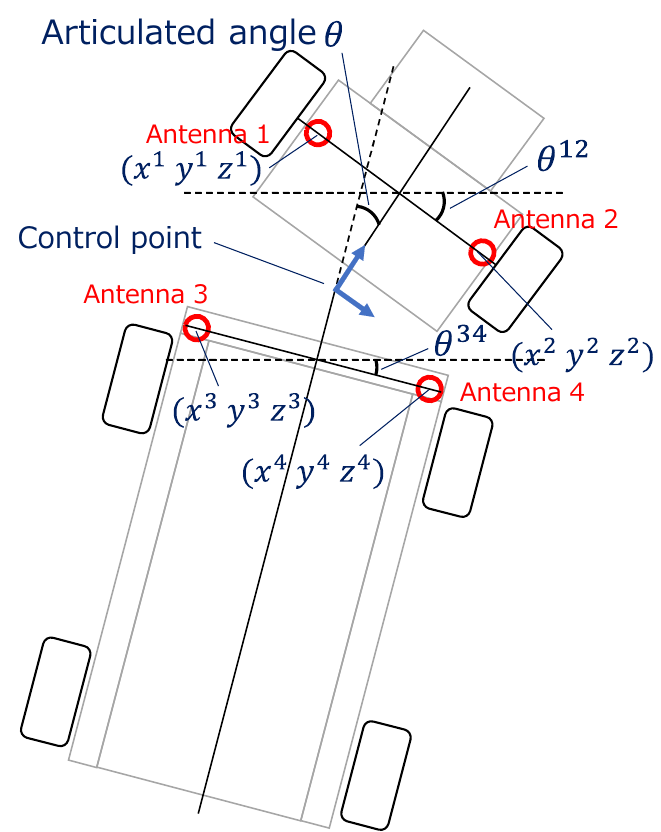} 
    \caption{Coordinate system and estimated state of the dump truck. The articulation angle can be calculated from the positions of the four GNSS antennas.}
    \label{fig:5}
\end{figure}

The orientation, articulation angle, and position of the dump truck are calculated from the 3D position of each antenna obtained by optimization. Figure 5 shows the coordinate system and articulation angle of the dump truck. The orientation angle is the angle of the front part of the articulated dump truck and is calculated using the following equation.

\begin{equation}
    {\Psi} = {\theta}_{\mathrm{12}} =\tan ^{-1}\left( \dfrac {y_{2}-y_{1}}{x_{2}-x_{1}}\right)
\end{equation}

\noindent The accuracy of the orientation angle depends on the length of the baseline vector between the antennas, but for the articulated dump truck developed here, the distance between the antennas is 2.8 m, and the orientation angle can be estimated with an accuracy of approximately 0.1 degree or less.

The next step is to calculate the articulation angle. The articulation angle is the difference in orientation between the front and rear of the dump truck and is calculated as follows.

\begin{equation}
    {\theta} = {\theta}_\mathrm{12} - {\theta}_\mathrm{34}  
     =\tan ^{-1}\left( \dfrac {y_{2}-y_{1}}{x_{2}-x_{1}}\right)
     -\tan ^{-1}\left( \dfrac {y_{4}-y_{3}}{x_{4}-x_{3}}\right)
\end{equation}

Finally, the position of the dump truck is calculated. The control point of the dump truck is the link position, and the link position is used to control the dump truck. Using the pre-measured position offset from the link position to each antenna, the dump truck position $\mathbf{x}_\mathrm{dump}$ can be estimated as follows.

\begin{equation}
    \mathbf{x}_\mathrm{dump} = \frac{
    \sum_{i=1}^2 \left( \mathbf{x}_i + \mathbf{R}(\theta_\mathrm{12}) \mathbf{B}_i \right)+ 
    \sum_{i=3}^4 \left( \mathbf{x}_i + \mathbf{R}(\theta_\mathrm{34}) \mathbf{B}_i \right)
    }{4}
\end{equation}

\noindent where, $\mathbf{R}(\cdot)$ is the rotation matrix and $\mathbf{B}_i$ is the pre-measured vector from each antenna to the dump truck control point. As described above, by averaging the dump truck positions calculated from each antenna, the accuracy of dump truck position estimation using CLAS positioning can be improved.

\section{Evaluation}
The accuracy of the proposed method in estimating orientation angle, articulation angle and position is evaluated in static and kinematic tests. As a comparison method, a method that does not use moving-base RTK-GNSS, but only uses CLAS positioning solutions is used. Furthermore, the method is compared to the conventional method of constraining the absolute position of a dump truck by RTK-GNSS using a GNSS reference station.

\begin{figure}[t!]
    \centering
    \includegraphics[width=170mm]{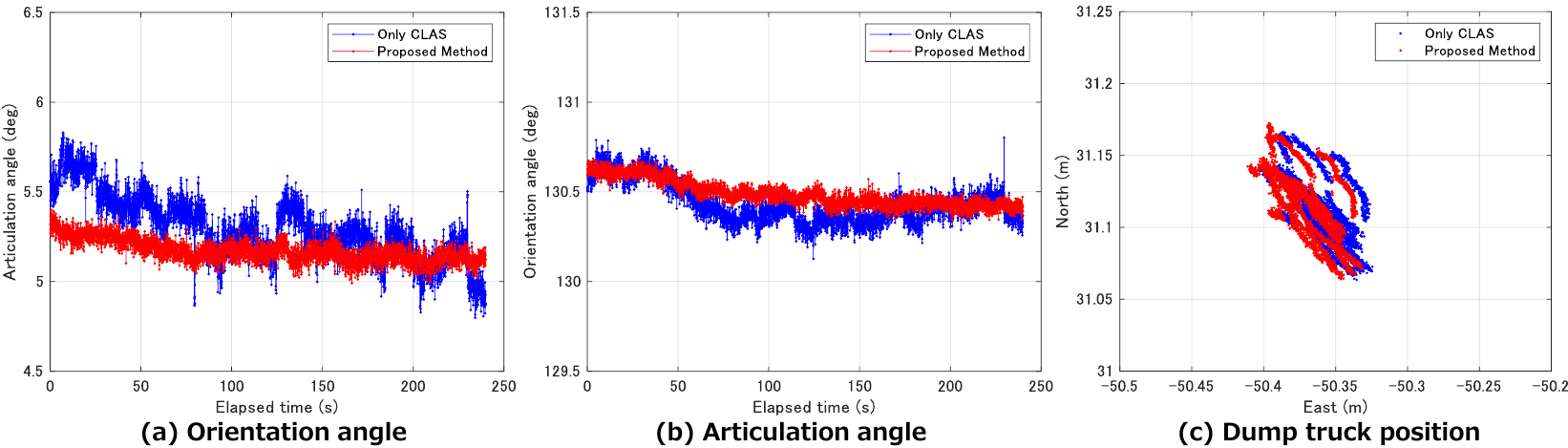} 
    \caption{Dump truck state estimation results for each method in the static test. (a) orientation angle, (b) articulation angle, and (c) dump truck position.}
    \label{fig:6}
\end{figure}
\begin{table}[t!]
    \centering
    \caption{Standard deviation of states estimated by each method in static tests.}
    \label{tab:1}
    \begin{tabular}{@{}ccccc@{}}
    \toprule
    \multicolumn{2}{c}{}                       & Only CLAS & Proposed Method & RTK-GNSS \\ \midrule
    \multicolumn{2}{c}{Orientation angle deg}  & 0.112     & 0.072           & 0.072    \\
    \multicolumn{2}{c}{Articulation angle deg} & 0.183     & 0.063           & 0.063    \\
    \multirow{3}{*}{Position m}    & East cm   & 1.775     & 1.818           & 0.145    \\
                                   & North cm  & 2.279     & 2.331           & 0.308    \\
                                   & Up cm     & 2.931     & 2.929           & 0.566    \\ \bottomrule
    \end{tabular}
\end{table}

\subsection{Static Experiment}
With a stationary dump truck, the orientation angle, articulation angle, and position are estimated by the comparison method and the proposed method, respectively, and the standard deviation of the estimated dump truck state is evaluated. The experimental environment was open-sky, and the dump truck was stationary for 5 minutes to estimate the orientation angle, articulation angle, and position. Figure 6 shows a 2D plot of the position estimated by each method and the time variation of the orientation angle and articulation angle. Table 1 shows the standard deviation of the estimated states by each method. When only the PPP-RTK solution of CLAS is used, it can be seen that the estimated orientation angle and articulation angle vary with the variation of the CLAS solution. However, when the proposed method is combined with moving-base RTK-GNSS, the estimation accuracy is improved. In the static test, the proposed method significantly improved from the CLAS-only method with respect to orientation and articulation angle estimation, but the position estimation accuracy remained almost the same. This is thought to be because the PPP-RTK positioning solution with CLAS contains similar bias errors in all antennas, so averaging them together did not improve the accuracy. The proposed method was able to estimate the standard deviation of the orientation angle by 0.072$^\circ$ and the standard deviation of the articulation angle by 0.063$^\circ$, with the same accuracy as the conventional method using GNSS reference stations.

\subsection{Kinematic Test}
A dump truck is driven to evaluate the orientation angle, articulation angle, and position estimation accuracy while driving the dump truck. Here, we compare the method using conventional RTK-GNSS with a GNSS reference station as a reference with the proposed method and the method using only CLAS-based positioning solutions. Figure 7 shows the trajectory of a dump truck during the experiment. As shown in Figure 7, the truck traveled the figure-8 route. The experimental environment was the same as in the static test, with an open-sky, and the dump truck traveled at a speed of approximately 10 km/h. Figure 8 shows the orientation angle error, articulation angle error, and 3D position error compared to the reference using RTK-GNSS with a GNSS reference station. Table 2 also shows the RMS error for each method. As can be seen from Figure 8 and Table 2, Compared to the static test, the CLAS positioning accuracy deteriorated in the kinematic test, and the estimation accuracy of orientation and articulation angle worsened. However, the proposed method, which combines moving-base RTK-GNSS, provided the same estimation accuracy as the conventional method using RTK-GNSS with a GNSS reference station. Regarding the accuracy of position estimation, the proposed method was also found to improve over the case of using CLAS alone.

\begin{figure}[t!]
    \centering
    \includegraphics[width=100mm]{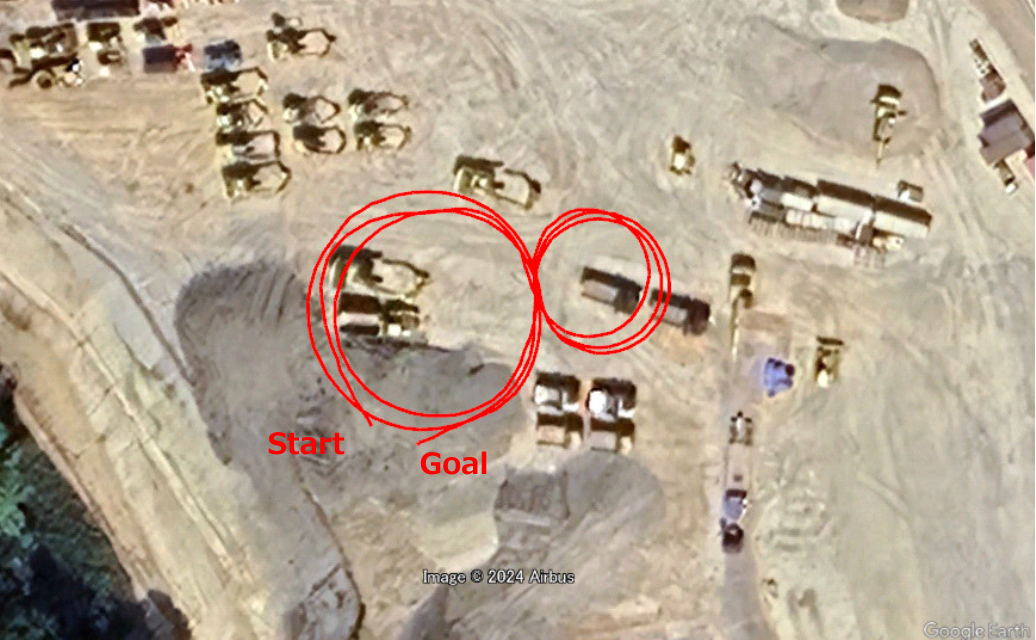} 
    \caption{Aerial photograph of the experimental field. The red line indicates the travel path of the articulated dump truck. The experimental field is an open-sky environment with few obstacles blocking the GNSS signals.}
    \label{fig:7}
\end{figure}

\begin{figure}[t!]
    \centering
    \includegraphics[width=170mm]{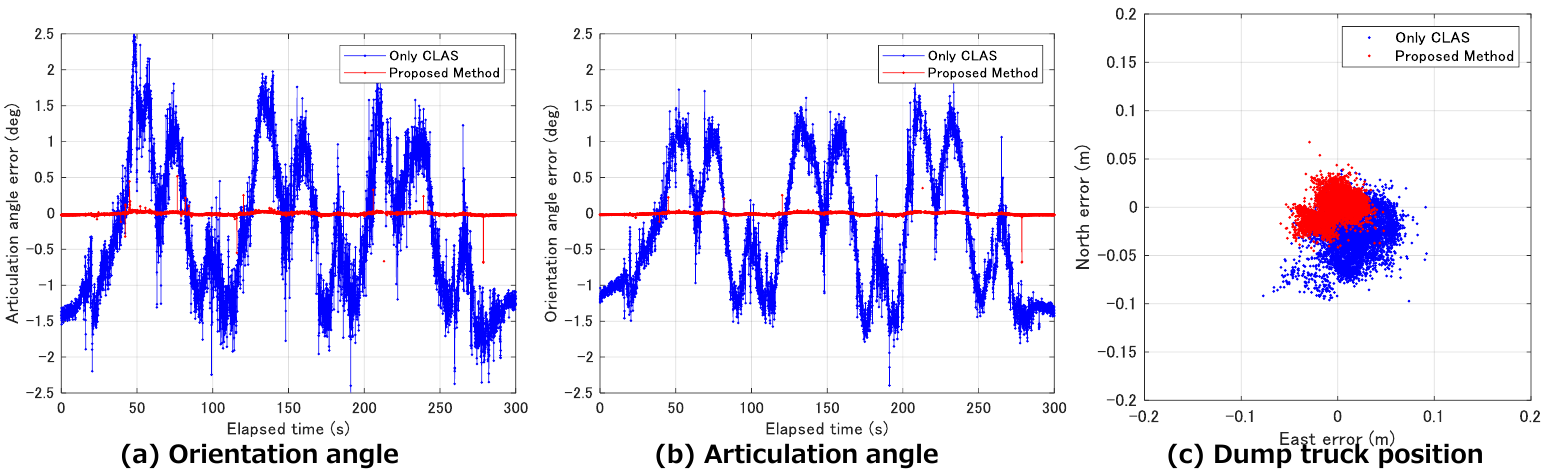} 
    \caption{Dump truck state estimation errors for each method using kinematic tests. (a) orientation angle , (b) articulation angle, and (c) dump truck position. Each method was compared with a conventional RTK-GNSS-based method using a GNSS reference station.}
    \label{fig:8}
\end{figure}

\begin{table}[t!]
    \centering
    \caption{RMS error of states estimated by each method in dynamic tests.}
    \label{tab:2}
    \begin{tabular}{@{}cccc@{}}
    \toprule
    \multicolumn{2}{c}{}                       & Only CLAS & Proposed Method \\ \midrule
    \multicolumn{2}{c}{Orientation angle deg}  & 0.936     & \textbf{0.021}  \\
    \multicolumn{2}{c}{Articulation angle deg} & 1.029     & \textbf{0.027}  \\
    \multirow{3}{*}{Position m}    & East cm   & 2.890     & \textbf{1.767}  \\
                                   & North cm  & 3.631     & \textbf{1.253}  \\
                                   & Up cm     & 2.856     & \textbf{2.937}  \\ \bottomrule
    \end{tabular}
\end{table}

Based on the above experiments, the proposed method compensates for the weakness of CLAS in QZSS by combining moving-base RTK-GNSS, and achieves highly accurate dump truck state estimation. The proposed method enables automatic operation of dump trucks without using ground communication or GNSS reference stations, and is expected to greatly expand the application environment for automatic operation of dump trucks.

\subsection{Conclusion}
This paper proposed a state estimation method of dump truck position, heading and articulation angle for automatic operation of articulated dump trucks. Four GNSS receivers and antennas are mounted on a dump truck, and the state of the dump truck is estimated from the geometric arrangement of the antennas. In addition to positioning each antenna using CLAS, we proposed a method to realize the state estimation of dump trucks by combining moving-base RTK-GNSS between antennas through factor graph optimization, which solves the problems of low ambiguity fix rate and low positioning accuracy of CLAS. The performance of the proposed method was evaluated in static and mobile tests in a real environment, and the results showed that the proposed method can estimate the state of a dump truck with the same level of accuracy as the conventional method using RTK-GNSS with a GNSS reference station. The performance of the proposed method is significantly better than that of the method using only CLAS.

\section*{acknowledgements}
This paper is based on results obtained from a project, 18065741, subsidized by the New Energy and Industrial Technology Development Organization (NEDO). 

\bibliographystyle{apalike}
\bibliography{IONPNT_2024}

\end{document}